\documentclass[11pt,a4paper]{article}
\usepackage[hyperref]{naaclhlt2018}
\usepackage{times}
\usepackage{latexsym}
\usepackage{booktabs}
\usepackage{tikz}
\usepackage{multirow}
\usepackage{pgfplots}
\usepackage{tikzscale}
\usepackage{url}
\usepackage[utf8]{inputenc}
\usepackage[russian,english]{babel}
\usepackage{CJKutf8}
\usepackage{footnote}
\usepackage{threeparttable}

\aclfinalcopy % Uncomment this line for the final submission

\usepackage{tablefootnote}

\usepackage{amssymb}
\usepackage{amsthm}
\usepackage{color}
\usepackage{amsmath}
\usepackage{graphicx}
\usepackage{multirow}
\usepackage{breqn}
\usepackage{url}
\usepackage{comment}
\usepackage{diagbox}

\usepackage{subcaption}
\usepackage{caption}
\usepackage{tikz}
\usepackage{verbatim}

\usepackage{mathrsfs}
\usepackage{verbatim}
\usepackage{times}
\usepackage{url}
\usepackage{latexsym}
\usepackage{color}
\usepackage{float}
\usepackage{array}
\usepackage{amsmath,amsfonts,amssymb}
\usepackage{algorithm}
\usepackage{algpseudocode}
\usepackage{graphicx}
\usepackage{subcaption}
\usepackage{booktabs}
\usepackage{multirow}
\usepackage{tikz,pgfplots}
\usepackage{pgfplotstable}
\usepackage{enumitem}
\usepackage{pifont}
\usepackage{fancyvrb}
\usepackage{footnote}
\makesavenoteenv{tabular}
\makesavenoteenv{table}
\usepackage{tikz-dependency}
\usepackage{array}
\newcolumntype{L}[1]{>{\raggedright\let\newline\\\arraybackslash\hspace{0pt}}m{#1}}
\newcolumntype{C}[1]{>{\centering\let\newline\\\arraybackslash\hspace{0pt}}m{#1}}
\newcolumntype{R}[1]{>{\raggedleft\let\newline\\\arraybackslash\hspace{0pt}}m{#1}}

\usetikzlibrary{arrows,shapes,positioning,shadows,trees}

\usepackage[font=small]{caption}

\usepackage{stmaryrd}
\newcommand{\mnote}[1]
{
}

\pgfplotscreateplotcyclelist{mark list*}{%
every mark/.append style={solid,fill=.!80!black},mark=*\\%
every mark/.append style={solid,fill=.!80!black},mark=square*\\%
every mark/.append style={solid,fill=.!80!black},mark=triangle*\\%
every mark/.append style={solid,fill=.!80!black},mark=halfsquare*\\%
every mark/.append style={solid,fill=.!80!black},mark=pentagon*\\%
every mark/.append style={solid,fill=.!80!black},mark=halfcircle*\\%
every mark/.append style={solid,fill=.!80!black,rotate=180},mark=halfdiamond*\\%
every mark/.append style={solid,fill=.!80!black!40},mark=otimes*\\%
every mark/.append style={solid,fill=.!80!black},mark=diamond*\\%
every mark/.append style={solid,fill=.!80!black},mark=halfsquare right*\\%
every mark/.append style={solid,fill=.!80!black},mark=halfsquare left*\\%
}
\pgfplotscreateplotcyclelist{color list}{%
red,blue,black,yellow,brown,teal,orange,violet,cyan,green!70!black,magenta,gray}
\pgfplotscreateplotcyclelist{linestyles}{solid,dashed,dotted}
\pgfplotscreateplotcyclelist{linestyles*}{solid,dashed,dotted,dashdotted,dashdotdotted}

\tikzset{every mark/.append style={scale=1.5}}

\pgfplotscreateplotcyclelist{mylist}{%
{blue,mark=*,solid},
{green!60!black,mark=triangle},
{brown!60!black,mark=diamond*},
{brown!60!black,mark options={fill=brown!40},mark=otimes*},
{yellow!60!black,mark=triangle*,mark options={fill=brown!40}},
{red,mark=square,}}

\newcommand{\supa}[1]{
}
\newcommand{\danr}[1]{
}

\enlargethispage{\baselineskip} % to add additional line to a page

\usepackage{xcolor}
\newcommand{\ensuretext}[1]{#1}

\newcommand{\xnmarker}{\ensuretext{\textcolor{blue}{\ensuremath{^{\textsc{X}}_{\textsc{N}}}}}}
\newcommand{\yvmarker}{\ensuretext{\textcolor{green}{\ensuremath{^{\textsc{Y}}_{\textsc{V}}}}}}

\newcommand{\mycomment}[3]{}

\newcommand{\xn}[1]{\mycomment{\xnmarker}{#1}{red}}
\newcommand{\yv}[1]{\mycomment{\yvmarker}{#1}{purple}}
\newcommand{\todo}[1]{\mycomment{\xnmarker}{#1}{red}}

\newcommand\numberthis{\addtocounter{equation}{1}\tag{\theequation}}

\newcommand{\balapinc}{\emph{BalAPinc }}
\newcommand{\slqs}{\emph{SLQS }}
\newcommand{\hard}{Hyper-Hypo}
\newcommand{\easy}{Hyper-Cohypo}

\newcommand{\ignore}[1]{}
\newcommand\blfootnote[1]{%
	\begingroup
	\renewcommand\thefootnote{}\footnote{#1}%
	\addtocounter{footnote}{-1}%
	\endgroup
}
\title{Robust Cross-lingual Hypernymy Detection using Dependency Context}

\author{Shyam Upadhyay$^{1*}$, Yogarshi Vyas$^{2*}$, Marine Carpuat$^{2}$, Dan Roth$^{1}$ \\
  $^{1}$ Department of Computer and Information Science,  University of Pennsylvania, Philadelphia, PA  \\
  $^{2}$ Department of Computer Science, University of Maryland, College Park, MD \\
  {\tt shyamupa@seas.upenn.edu, yogarshi@cs.umd.edu, }\\ 
  {\tt marine@cs.umd.edu, danroth@seas.upenn.edu} \\}

\date{}

\begin{document}
\maketitle

\begin{abstract}
{\em Cross-lingual Hypernymy Detection}\blfootnote{$^*$ These authors contributed equally.} involves determining if a word in one language (``fruit'') is a hypernym of a word in another language (``pomme'' i.e. apple in French). 
The ability to detect hypernymy cross-lingually can aid in solving cross-lingual versions of tasks such as textual entailment and event coreference.
We propose {\sc BiSparse-Dep}, a family of unsupervised approaches for cross-lingual hypernymy detection, which learns sparse, bilingual word embeddings based on dependency contexts. We show that {\sc BiSparse-Dep} can significantly improve performance on this task, compared to approaches based only on lexical context. Our approach is also robust, showing promise for low-resource settings: our dependency-based embeddings can be learned using a parser trained on related languages, with negligible loss in performance. We also crowd-source a challenging dataset for this task on four languages -- Russian, French, Arabic and Chinese. Our embeddings and datasets are publicly available.\footnote{\url{https://github.com/yogarshi/bisparse-dep/}}

%%% Local Variables:
%%% mode: latex
%%% TeX-master: "main"
%%% TeX-PDF-mode: t
%%% End: 

\end{abstract}

\section{Introduction}
\label{sec:intro}
%%\mc{Marine comments on intro : 1) Why is xlingual hypernymy detection particularly
%%	imp. in low resource settings? 2) Abstract + intro still makes low resource sound
%%	like a side note 3) What is the core theme? Something like robust unsupervised models
%	for comparing meaning beyond simialrity/translation equivalence?}

% Why is hypernymy detection important
% The ability to identify semantic relationships between lexical items 
% is necessary for \ignore{sophisticated} multiple natural
% language understanding tasks such as Question Answering
% (QA)~\cite{harabagiu2006methods}, Recognizing Textual Entailment
% (RTE)~\cite{dagan2013recognizing}, and Knowledge
% Acquisition~\cite{romano2006investigating}. One such important relationship is \emph{hypernymy}.
%  For instance, knowing that \emph{game} is a hypernym of \emph{chess}, can help a QA 
% system answer the question ``Which game does Magnus Carlsen play?"

% Why cross-lingual hypernymy is important
Translation helps identify correspondences in bilingual texts, but other asymmetric
semantic relationships can improve language understanding when translations are not exactly equivalent. One such relationship is \emph{cross-lingual hypernymy} -- 
 identifying that {\em écureuil} (``squirrel" in French)
is a kind of {\em rodent}, or 
\foreignlanguage{russian}{ворона}
(``crow" in Russian) 
is a kind of {\em bird}. The ability to detect hypernyms across languages serves as a building block in a range of
cross-lingual tasks, including Recognizing Textual Entailment (RTE)~\cite{Negri2012,Negri2013a}, constructing multilingual taxonomies~\cite{fu-EtAl2014}, event
coreference across multilingual news sources~\cite{vossen2015interoperability}, and evaluating Machine Translation output~\cite{PGJM09}.

% Challenges in cross-lingual entailment
Building models that can robustly identify hypernymy 
across the spectrum of human languages is a challenging 
problem, that is further compounded in low resource settings. 
At first glance, translating words to English and then identifying hypernyms in a monolingual setting
may appear to be a sufficient solution. However, this approach cannot capture many phenomena.
For instance, the English words {\em cook}, {\em leader} and {\em
  supervisor} can all be hypernyms of the French word {\em chef}, as the French word
does not have a exact translation in English covering its possible
usages. However, translating {\em chef} to {\em cook} and then determining hypernymy monolingually
precludes identifying {\em leader} or {\em supervisor} as a hypernyms of {\em chef}.
Similarly, language-specific usage patterns can also influence
hypernymy decisions. For instance, the French word {\em chroniqueur}
translates to {\em chronicler} in English, but is more frequently used
in French to refer to journalists (making {\em journalist} its hypernym).\footnote{All examples are from our dataset.}
% For instance, the English word \emph{drug} can be translated into
% French as \textit{drogue} (``narcotic") or \textit{médicament}
% (``medicine"), and depending on how it is translated, it may or may
% not be a hypernym of the French word \textit{aspirine} (``aspirin").

This motivates approaches that {\em directly} detect hypernymy in the cross-lingual setting by extending distributional methods for
detecting monolingual hypernymy, as in our prior work 
\cite{Vyas2016}. State-of-the-art
distributional approaches~\cite{Roller2016,shwartz2017hypernymy} for detecting
monolingual hypernymy require syntactic analysis (eg. dependency
parsing), which may not available for many
languages. Additionally, limited training resources make unsupervised methods more desirable than supervised hypernymy detection approaches~\cite{Roller2016}\xn{add more citations here}.
Furthermore, monolingual distributional approaches cannot be applied
directly to the cross-lingual task, because the vector spaces of two
languages need to be aligned using a cross-lingual resource (a bilingual dictionary, for instance).

We tackle these challenges by proposing {\sc BiSparse-Dep} - 
a family of robust, unsupervised approaches for identifying cross-lingual hypernymy.
{\sc BiSparse-Dep} uses a cross-lingual word embedding model learned from a 
small bilingual dictionary and a variety of monolingual syntactic context extracted 
from a dependency parsed corpus.
% attractions
{\sc BiSparse-Dep} exhibits robust behavior along multiple dimensions.
In the absence of a dependency treebank for a language, it
can learn embeddings using a {parser} trained on related
languages. When exposed to less monolingual data, or a lower quality
bilingual dictionary, {\sc BiSparse-Dep} degrades only marginally. In
all these cases, it compares favorably with models that have been
supplied with all necessary resources, showing promise for
low-resource settings. We extensively evaluate {\sc BiSparse-Dep} on
a new crowd-sourced cross-lingual dataset, with over 2900 hypernym
pairs, spanning four languages from distinct families -- French, Russian, Arabic and Chinese -- and release the datasets for
future evaluations.

% Cross-lingual lexical entailment helps in obtaining a more refined understanding of semantic mappings across languages that is more concrete than semantic similarity.

% Such information can be useful in many multilingual applications,
% including machine translation and its evaluation~\cite{PGJM09}, question
% answering~\cite{harabagiu2006methods}, or RTE \cite{Negri2012,Negri2013a}. In particular, cross-lingual hypernymy detection
% can be applied for tasks requiring knowledge consolidation across
% languages, like constructing taxonomies~\cite{fu-EtAl2014}, or event coreference~\cite{vossen2015interoperability}.

% resolution~\cite{martins2015transferring}, where entailment can help
% identify nominal coreferring mentions.  \todo{Maybe elaborate a bit more here, or give example}

\ignore{
	\todo{this should be echoed in the intro somehow}
	Note that relying on translations for this task may not suffice, as
	translation need not preserve the entailment relationships due to various
	reasons such as translation ambiguity, noisy translation, etc \yv{We
		probably need an example here}. Furthermore, lack of cross-lingually
	annotated training data motivates developing unsupervised approaches
	for detecting entailment, which have been well-studied in monolingual
	settings \cite{}. % To port such techniques to the cross-lingual
	% settings, \newcite{Vyas2016} motivate the use of sparse, bilingual
	% word representations.
	\supa{Say precisely what we need to construct a taxonomy/other useful
		resource, and why is it atractive. Say language indepdencent}
}

% \yv{Stuff moved from motivation - END}

%%% Local Variables:
%%% mode: latex
%%% TeX-master: "main"
%%% TeX-PDF-mode: t
%%% End: 

\section{Related Work}
\label{sec:related}
\begin{figure*}[!ht]
  \centering
  \includegraphics[scale=0.45]{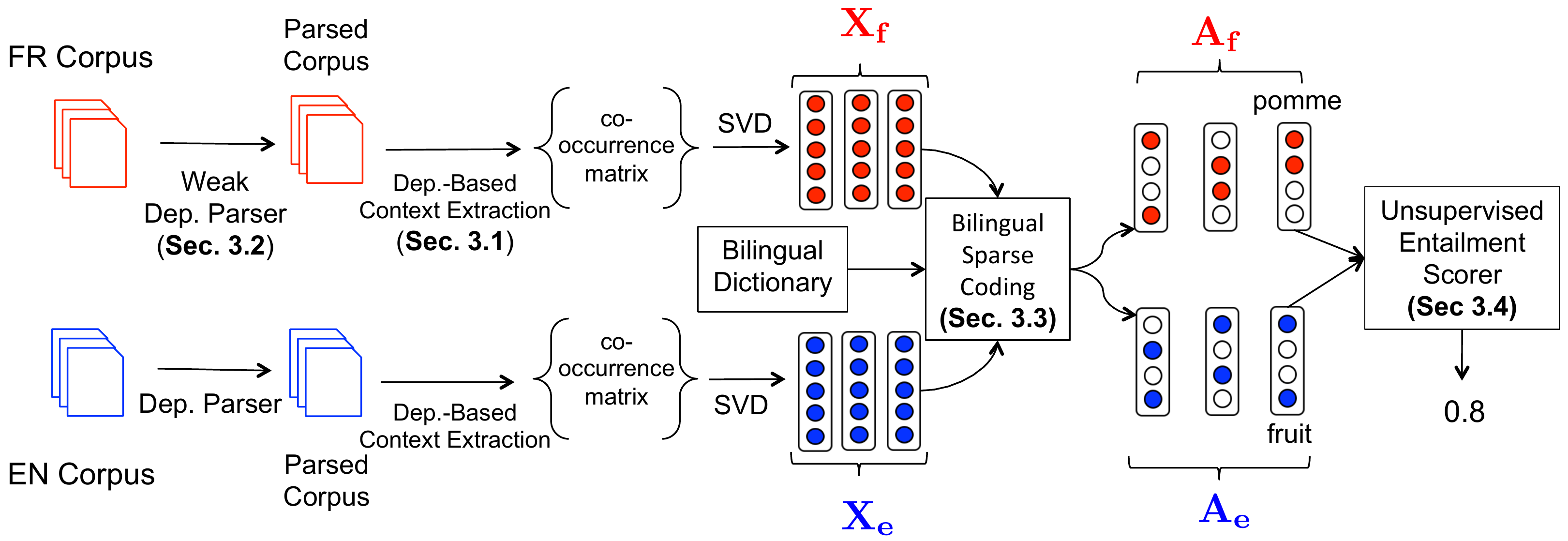}
  \caption{The {\sc BiSparse-Dep} approach, which learns sparse bilingual embeddings using dependency based contexts. The resulting sparse embeddings, together with an unsupervised entailment scorer, can detect hypernyms across languages (e.g., {\em pomme} is a {\em fruit}).}
  \label{fig:schema}
  \vspace{-0.15in}
\end{figure*}
%%% Local Variables:
%%% mode: latex
%%% TeX-master: "main"
%%% TeX-PDF-mode: t
%%% End: 

%Our work is broadly related to the following areas of research.
\paragraph{Cross-lingual Distributional Semantics}
Cross-lingual word embeddings have been shown to encode semantics across languages in
tasks such as word similarity~\cite{faruqui:2014} and lexicon
induction~\cite{vulic:2015}.
% \newcite{Upadhyay2016}, \newcite{levy-sogaard-goldberg2017}, 
% and \newcite{Vulic2016} discuss
% the impact of various training methods and data on a spectrum of
% tasks. 
Our works stands apart in two aspects (1) In contrast to tasks
involving similarity and synonymy (symmetric relations), the focus of
our work is on detecting {\em asymmetric} relations across languages,
using cross-lingual embeddings. (2) Unlike most previous work, we use
dependency context instead of lexical context to induce cross-lingual
embeddings, which allows us to abstract away from language specific
word order, and (as we show) improves hypernymy detection.

More closely related is our prior work \cite{Vyas2016} where we used lexical
context based embeddings to detect cross-lingual lexical
entailment. In contrast, the focus of this work is on hypernymy, a more well-defined
relation than entailment. Also, we improve upon our previous approach
by using dependency based embeddings
(\S\ref{sec:dependency-vs-window}), and show that the improvements
 hold even when exposed to data scarce settings
(\S\ref{sec:eval-robustn-sc}). We also do a more comprehensive
evaluation on four languages paired with English, instead of just French.

\paragraph{Dependency Based Embeddings} 
In monolingual settings, dependency based embeddings have been shown
to outperform window based
embeddings on many
tasks~\cite{bansal:2014,simlex,Melamud2016}. \newcite{Roller2016} showed that dependency embeddings can help
in recovering Hearst patterns~\cite{Hearst92} like ``animals such as
cats'', which are known to be indicative of hypernymy. \newcite{shwartz2017hypernymy} demonstrated that
dependency based embeddings are almost always superior to window based
embeddings for identifying hypernyms in English. Our work
uses dependency based embeddings in a cross-lingual setting, a less explored research direction. A key
novelty of our work also lies in its use of syntactic transfer to derive dependency contexts. This
scenario is more relevant in a cross-lingual setting, where treebanks
might not be available for many languages.

\section{Our Approach -- {\sc BiSparse-Dep}}
\label{sec:depend-based-word}
\begin{figure}[!t]
    \centering
    \scalebox{.9}{\input{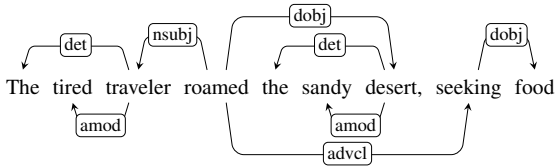}}
    \caption{Example Dependency Tree.}
    
    \label{fig:deptree}
\end{figure}

% \begin{figure}
  % \begin{subfigure}[b]{0.45\linewidth}
    % \centering
    % \input{tree}
    % \caption{Example Dependency Tree.}
    % \label{fig:deptree}
    %% \vspace{2ex}
  % \end{subfigure}
  % \hfill
  % \begin{subfigure}[b]{0.45\linewidth}
  %   \centering
  %   \footnotesize
  %   \begin{tabular}{cc}
  %     Type & Context of {\em traveler}\\
  %     \toprule
  %     {\sc Full} & roamed\#nsubj$^{-1}$, tired\#amod\\
  %     {\sc Joint} & roamed\#desert, roamed\#searching\\ 
  %     % {\sc Unlabeled} & roamed, tired\\ 
  %     \bottomrule
  %   \end{tabular}
  %   \caption{Different syntactic contexts of {\em traveler}.}
  %   \label{fig:labels}
  % \end{subfigure}
  % \caption{Example dependency tree and different syntactic contexts described using {\sc Full}, {\sc Joint} and {\sc Unlabeled}}
%   \label{fig:depcxt}
% \end{figure}

%%% Local Variables:
%%% mode: latex
%%% TeX-master: "main"
%%% TeX-PDF-mode: t
%%% End: 

We propose \textsc{BiSparse-Dep},
a family of approaches that uses sparse, bilingual, dependency based word
embeddings to identify cross-lingual hypernymy.

Figure~\ref{fig:schema} shows an overview of the end-to-end pipeline of {\sc BiSparse-Dep}. The two key components of this pipeline are: (1) {\em
  Dependency based contexts} (\S\ref{sec:depend-based-word-1}), which
help us generalize across languages with minimal customization by
abstracting away language-specific word order. We also discuss how to
extract such contexts in the absence of a treebank in the
language (\S\ref{sec:depend-embedd-low}) using a (weak) dependency
parser trained on related languages. (2) {\em Bilingual sparse coding} (\S\ref{subsec:bisparse}),
which allows us to align dependency based word embeddings in a
shared semantic space using a small bilingual dictionary. The
resulting sparse bilingual embeddings can then be used with a
unsupervised entailment scorer (\S\ref{sec:unsup-enta-scor}) to predict hypernymy for cross-lingual word pairs.

% \ignore{Dependency based embeddings have been shown to capture
% functional similarity (eg. {\em singing} and {\em rapping}), in contrast to topical
% similarity (eg. {\em singing} and {\em dancing})~\cite{levygoldberg2014}, and have hence proven to be useful for monolingual hypernymy detection \cite{shwartz2017hypernymy}.} % Here, we investigate the applicability of the same to the cross-lingual version of the task. 

\subsection{Dependency Based Context Extraction}
\label{sec:depend-based-word-1}

The context of a word can be described in multiple ways using its syntactic
neighborhood in a dependency graph. For instance, in
Figure~\ref{fig:deptree}, we describe the context for a target word
({\em traveler}) in the following two ways:

\begin{itemize}[noitemsep]
  \setlength\itemsep{0em}
\item {\sc Full} context ~\cite{Pado2007DCS,Baroni2010,levygoldberg2014}: Children and parent words, concatenated with the label and direction of the relation (eg. {\em roamed\#nsubj}$^{-1}$ and {\em tired\#amod} are contexts for {\em traveler}).
\item {\sc Joint}~ context \cite{chersoniEMNLP2016}: Parent
  concatenated with each of its siblings (eg. {\em roamed\#desert} and {\em roamed\#seeking} are contexts for {\em traveler}).
\end{itemize}
These two contexts exploit different amounts of syntactic
information -- {\sc Joint} does not require labeled parses, unlike {\sc
  Full}. The {\sc Joint} context combines parent
and sibling information, while {\sc Full} keeps them as distinct
contexts. Both encode directionality into the context, either through label direction or through sibling-parent relations.
% Also note that the co-occurrence matrix generated using these
% contexts (except {\sc Unlabeled}) is asymmetric.

We use word-context co-occurrences generated using these 
contexts in a distributional semantic model (DSM) in
lieu of window based contexts to generate dependency based embeddings.

% \ignore{Traditionally, distributional semantic models (DSMs) have been constructed using
% window based contexts, where a word-context co-occurrence matrix is built by collecting counts between target words and adjacent words within a fixed window
% around it~\cite[{\em inter alia}]{levy2001learning,mikolov2013efficient,Pennington2014}.}
%

\subsection{Dependency Contexts without a Treebank}
\label{sec:depend-embedd-low}
Using dependency contexts in multilingual settings may not always be
possible, as dependency treebanks are not available for many
languages. To circumvent this issue, we use related languages to train a
weak dependency parser.

We train a {\em delexicalized} parser using treebanks of related
languages, where the word form based features are turned off, so that
the parser is trained on purely non-lexical features (e.g. POS
tags). The rationale behind this is that related languages
show common syntactic
structure that can be transferred to the original language, with
delexicalized parsing~\cite[\em inter
alia]{zeman2008cross,mcdonald2011multi} being one popular
approach.\footnote{More sophisticated techniques for
transferring syntactic knowledge have been
proposed~\cite{TACL892,rasooliTACL}, but we prioritize simplicity and show that a
simple delexicalized parser is effective.}

\subsection{Bilingual Sparse Coding}
\label{subsec:bisparse}
 Given a dependency based 
co-occurrence matrix described in the previous section(s),
we generate {\sc BiSparse-Dep}
embeddings using the framework from our prior work \cite{Vyas2016}, which we
henceforth call \textsc{BiSparse}. \textsc{BiSparse} generates sparse,
bilingual word embeddings using a dictionary learning objective with a
sparsity inducing $l_1$ penalty. We give a brief overview of this
approach, the full details of which can be found in our prior work.

For two languages with vocabularies $v_e$ and $v_f$, and monolingual dependency embeddings $\mathbf{X_e}$ and $\mathbf{X_f}$, {\sc BiSparse} solves the following objective:
\begin{align*}
	\label{eqn:bisparse}
	\underset{\mathbf{A_e, D_e, A_f, D_f}}{\text{argmin~}}& \sum\limits_{i=1}^{v_e}  \; \frac{1}{2}|| \mathbf{A_e}_{i}\mathbf{D_e}^{\text{T}} - \mathbf{X_e}_i  ||^2_2 \; + \lambda_e||\mathbf{A_e}_i||_1 \;\\                                
	+& \sum\limits_{j=1}^{v_f}  \; \frac{1}{2}|| \mathbf{A_f}_j\mathbf{D_f}^{\text{T}} - \mathbf{X_f}_j ||^2_2 \; + \lambda_f||\mathbf{A_f}_j||_1 \; \\          
	+& \sum\limits_{i,j} \frac{1}{2} \lambda_x \mathbf{S}	_{ij} || \mathbf{A_e}_i - \mathbf{A_f}_j ||_2^2 \numberthis\\  
	{\text{s.t.~}} & \mathbf{A_{k}} > \mathbf{0} \qquad \|{\mathbf{D_{k}}_i}\|_2^2\leq 1 \qquad \mathbf{k} \in \{\mathbf{e},\mathbf{f}\}
\end{align*}
\ignore{The first two rows and the constraints in Equation \ref{eqn:bisparse} encourage sparsity and non-negativity in the embeddings, by solving a sparse coding problem where ($\mathbf{D_e}$, $\mathbf{D_f}$) represent the dictionary matrices and ($\mathbf{A_{e}}, \mathbf{A_{f}}$) the final sparse representations. The third row imposes bilingual constraints, weighted by the regularizer $\lambda_x$, so that words that are strongly aligned according to $\mathbf{S}$ have similar representations. The above non-convex optimization problem can be solved using a proximal gradient method such as FASTA ~\cite{Goldstein2014}.}
where $\mathbf{S}$ is
a translation matrix, and  
$\mathbf{A_{e}}$ and $\mathbf{A_{f}}$ are sparse matrices which are bilingual
representations in a shared semantic space. The translation matrix
$\mathbf{S}$ (of size $v_e \times v_f$) captures correspondences
between the vocabularies (of size $v_e$ and $v_f$) of two
languages. For instance, each row of $\mathbf{S}$ can be a one-hot
vector that identifies the word in \textbf{\emph{f}} that is most
frequently aligned with the \textbf{\emph{e}} word for that row in a
large parallel corpus, thus building a one-to-many mapping between the
two languages.

\subsection{Unsupervised Entailment Scorer}
\label{sec:unsup-enta-scor}
% \textbf{Unsupervised Entailment Metrics.}
A variety of scorers can be used to quantify the
directional relationship between two words, given feature representations
of these words
\cite{lin1998automatic,weeds2003general,lenci2012identifying}.
Once the \textsc{BiSparse-Dep} embeddings are constructed, we
use \balapinc\cite{kotlerman2009directional} to score word pairs for 
hypernymy. \balapinc is
based on the distributional inclusion hypothesis~\cite{Geffet2005}
and computes the geometric mean
of 1) \emph{LIN} \cite{lin1998automatic}, a symmetric score that
captures similarity, and 2)
\emph{APinc}, an asymmetric score based on average precision.
%, for a word pair ($u$,$v$).

% Thus,
% \begin{align*}
% \text{\em BalAPinc}(u \rightarrow v) = \sqrt{\text{\emph{LIN}}(u,v)\cdot\text{\emph{APinc}}(u \rightarrow v)}
% \end{align*}

%%% Local Variables:
%%% mode: latex
%%% TeX-master: "main"
%%% TeX-PDF-mode: t
%%% End: 

% \section{Learning Sparse Cross-Lingual Embeddings}
% \label{sec:learn-sparse-cross}
% \input{models}

\section{Crowd-Sourcing Annotations}
\label{sec:evaluation-dataset}
There is no publicly available dataset to evaluate models of hypernymy detection across multiple languages. 
 \ignore{As identified by \newcite{carmona2017well},  it is crucial to carefully design a dataset to make consistent claims about a hypernymy detection model.} 
\yv{I  removed the citation to the carmona and riedel paper, since their paper actually makes claims only about supvervised hypernymy detection.}
While ontologies like  Open Multilingual WordNet
(OMW)~\cite{bond-foster:2013:ACL2013} and
BabelNet~\cite{NavigliPonzetto:12aij} contain cross-lingual links, these
resources are semi-automatically generated and hence contain 
noisy edges. Thus, to get reliable and high-quality test beds, 
we collect evaluation datasets using CrowdFlower\footnote{\url{http://crowdflower.com}}. Our datasets span four languages from distinct
families - French (Fr), Russian (Ru), Arabic (Ar) and Chinese
(Zh) - paired with English. 
% Additionally, while \newcite{Vyas2016} only collected positive instances via
% crowdsourcing, we improve on their annotation protocol to collect
% negative examples as well.
%
% \ignore{ which are a Romance language, an East Slavic language, a
%   Central Semitic language, and a Sino-Tibetan language respectively.}
% \ignore{ \todo{Someone might say generality would have been shown if
%     we used different SVO orders. Do we want to safeguard against it?}
% }

To begin the annotation process, we first pool candidate pairs using
hypernymy edges across languages from OMW and BabelNet, along with
translations from monolingual hypernymy
datasets~\cite{Baroni2011,Baroni:2012:EAW:2380816.2380822,kotlerman2010directional}.

\subsection{Annotation Setup}
\label{subsec:annsetup}
% \todo{Should we add screenshots of annotation process to appendix?}
The annotation task requires annotators to be fluent in 
both English and the non-English language. To ensure only fluent
speakers perform the task, for each language, we provide task
instructions in the non-English language itself. Also, we restrict the task to
annotators verified by CrowdFlower to have those language skills.
% These two steps are meant to prevent non-fluent speakers from
% performing the task.  Third, examples are shown to the users before
% they perform the task.
Finally, annotators also need to pass a quiz based on a small amount
of gold standard data to gain access to the task.

Annotators choose
between three options for each word pair ($p_{f},q_{e}$), where $p_f$ is a non-English word and $q_e$ is a English word : ``{$p_{f}$ {is a kind of} $q_{e}$}'',
``{$q_{e}$ is a part of $p_{f}$}'' and ``none of the above''.  Word
pairs labeled with the first option are considered as positive
examples while those labeled as ``none of the above'' are considered
as negative.\footnote{We collected more negative pairs than positive,
  but sampled so as to keep a balanced dataset for ease of
  evaluation. We will release all annotated pairs along with the
  dataset.} The second option was included to filter out meronymy
examples that were part of the noisy pool. We leave it to the annotator to infer whether the relation holds between any senses of $p_f$ or $q_e$, if either of them are polysemous.

For every candidate hypernym pair $(p_{f},q_{e})$, we also ask annotators to judge 
its reversed and translated \textit{hyponym} pair
$(q_{f},p_{e})$. 
% An annotator is shown a candidate hypernym pair $(A_{f},B_{e})$, along
% with the reversed and translated \textit{hyponym} pair
% $(b_{f},a_{e})$. 
For instance, if $(citron,food)$ is a hypernym candidate,
we also show annotators $(aliments, lemon)$ which is a
potential hyponym candidate (\textit{potential}, because as mentioned in \S\ref{sec:intro}, translation need not preserve semantic relationships).  The purpose of presenting the hyponym pair,
$(q_{f},p_{e}),$ is two-fold.  First, it emphasizes the directional
nature of the task. Second, it identifies
hyponym pairs, which we use as negative examples. The
hyponym pairs are challenging since
differentiating them from hypernyms truly requires detecting
asymmetry.

\begin{table}[t]
	\footnotesize
	\centering
	\begin{tabular}{lC{2.5cm}C{1.8cm}}
		\toprule
		% \multirow{2}{*}{Pair} & \#crowdsourced & \#pos (= \#neg) & \multirow{2}{*}{Fleiss $\kappa$}\\
		{pair} & \#crowdsourced & \#pos (= \#neg)\\
		% & & & \\
		\midrule
		French-English & 2115 & 763 \\           % 0.580508
		Russian-English & 2264 & 706 \\           % 0.536997
		Arabic-English & 2144 & 691 \\           % 0.532270
		Chinese-English & 2165 & 806 \\           % 0.557932
		\bottomrule
	\end{tabular}
	\caption{Crowd-sourced dataset statistics. \#pos (\#neg) denote positives (negatives) in the evaluation set. We deliberately under-sample negatives to have a balanced evaluation set.}
	\label{tab:dataset}
\end{table}

% \begin{table}[t]
% 	\footnotesize
% 	\centering
% 	\begin{tabular}{lp{1.8cm}p{1.8cm}c}
% 		\toprule
% 		% \multirow{2}{*}{Pair} & \#crowdsourced & \#pos (= \#neg) & \multirow{2}{*}{Fleiss $\kappa$}\\
% 		{pair} & \#crowdsourced & \#pos (= \#neg) & {Fleiss $\kappa$}\\
% 		% & & & \\
% 		\midrule
% 		Fr-En & 2115 & 763 & 58.1 \\           % 0.580508
% 		Ru-En & 2264 & 706 & 53.7 \\           % 0.536997
% 		Ar-En & 2144 & 691 & 53.2 \\           % 0.532270
% 		Zh-En & 2165 & 806 & 55.8 \\           % 0.557932
% 		\bottomrule
% 	\end{tabular}
% 	\caption{Crowd-sourced dataset statistics. \#pos (\#neg) denote positives (negatives) in the evaluation set. We deliberately under-sample negatives to have a balanced evaluation set.}
% 	\label{tab:dataset}
% \end{table}

Each pair was judged by at least 5 annotators, and judgments
with 80\% agreement (at least 4 annotators agree) are considered
for the final dataset. This is a stricter condition than certain
monolingual hypernymy datasets - for instance, EVALution
\cite{santus2015evalution} - where agreement by 3 annotators is deemed
sufficient. Inter-annotator agreement measured using Fleiss'
Kappa~\cite{fleiss1971measuring} was 58.1 (French), 53.7 (Russian), 53.2 (Arabic)
and 55.8 (Chinese). This indicates moderate agreement, on par with
agreement obtained on related fine-grained semantic
tasks~\cite{Pavlick2015a}. We cannot compare with monolingual
hypernymy annotator agreement as, to the best of our knowledge, such
numbers are not available for existing test sets. Dataset
statistics are shown in Table~\ref{tab:dataset}.

We observed that annotators were able to agree on pairs containing
polysemous words where hypernymy holds for some
sense. For instance, for the French-English pair ({\em avocat}, {\em
  professional}), the French word {\em avocat} can either mean 
\textit{lawyer} or \textit{avocado}, but the pair was annotated as a
positive example. Hence, we leave it to the annotators to handle
polysemy by choosing the most appropriate sense.
% We note that that related work on monolingual hypernymy have not
% reported such an analysis, instead relying on \%age agreement.
% Although we present candidate pairs which are potential meronyms, meronyms make up $<$ 10\% of positive examples in our final dataset.
\subsection{Two Evaluation Test Sets}
\label{subsec:testsplits}
To verify if the crowdsourced hyponyms are challenging negative examples we create two
evaluation sets. Both share the (crowdsourced) positive
examples, but differ in their negatives:
\begin{itemize}[noitemsep]
	\item {\sc \hard} -- negative examples are the crowdsourced hyponyms.
	\item {\sc \easy} -- negative examples are {\em cohyponyms}
          drawn from OMW.
\end{itemize} 
Cohyponyms are words sharing a common hypernym. For instance, {\em bi\`ere}
(``beer" in French) and {\em vodka} are cohyponyms since they share a
common hypernym in \textit{alcool}/\textit{alcohol}. We choose cohyponyms for the
second test set because: (a) They require differentiating
between similarity (a symmetric relation) and hypernymy (an asymmetric relation). For instance, {\em bi\`ere} and {\em
  vodka} are highly similar yet, they do not have a hypernymy
relationship. (b) Cohyponyms are a popular choice of negative examples in many entailment datasets~\cite{Baroni2011}.
% Cohyponymy is symmetric in nature and
% hence, \todo{if its potentially easier, then what's the point of doing the experiment?}potentially easier to detect than hyponymy, which is
% asymmetric.

%%% Local Variables:
%%% mode: latex
%%% TeX-master: "main"
%%% TeX-PDF-mode: t
%%% End: 

\section{Experimental Setup}
\label{sec:exp-setup}
\subsection{Data and Evaluation Setup}
Training {\sc BiSparse-Dep} requires a dependency parsed monolingual
corpus, and a translation matrix for jointly
aligning the monolingual vectors.
% As described in \S\ref{subsec:bisparse}, our approach requires
% monolingually trained dependency vectors and a translation matrix for
% inducing the sparse bilingual vectors. 
We compute the translation matrix using word alignments derived from
parallel corpora (see corpus statistics in Table~\ref{tab:datastats}). While we use parallel corpora to generate the translation matrix to be 
comparable to baselines (\S\ref{sec:contr-appr}), we can obtain the
matrix from any bilingual dictionary.

The monolingual corpora are parsed using {\tt Yara
	Parser}~\cite{RasooliT15}, trained on the corresponding treebank
from the Universal Dependency Treebank~\cite{UnivDep} (UDT-v1.4).
% Parser choice was driven by parsing speed, and {\tt Yara
%  Parser} was shown to be fast and competitive with state-of-the-art
%  parsers~\cite{choi2015depends}.
{\tt Yara Parser} was chosen as it is fast, and competitive with
state-of-the-art parsers~\cite{choi2015depends}.  The monolingual
corpora was POS-tagged using
TurboTagger~\cite{martins-almeida-smith2013}.
\ignore{ We used the Stanford Segmenter for Chinese and
	Arabic~\cite{I05-3027,monroe-green-manning2014}.  } We induce dependency contexts for words by
first thresholding the language vocabulary to the top 50,000 
nouns, verbs and adjectives. A co-occurrence matrix is
computed over this vocabulary using the context types in
\S\ref{sec:depend-based-word-1}.

\paragraph{Inducing Dependency Contexts} The entries of the word-context co-occurrence matrix are re-weighted
using Positive Pointwise Mutual
Information~\cite{Bullinaria07extractingsemantic}. The resulting matrix is reduced
 to 1000 dimensions using SVD~\cite{golub1965calculating}.\footnote{Chosen based on preliminary experiments with \{500,1000,2000,3000\} dimensional vectors for En-Fr.} These vectors are used as $\mathbf{X_{e}},\mathbf{X_{f}}$ in the setup from \S\ref{subsec:bisparse} to generate 100 dimensional sparse bilingual vectors.

\paragraph{Evaluation}
We use accuracy as our evaluation metric, as it is easy to
interpret when the classes are balanced~\cite{turney2015experiments}.
Both evaluation datasets -- {\sc \hard}
and {\sc \easy} -- are split into 1:2 dev/test splits. 
% We use \balapinc~\cite{kotlerman2009directional} in all our
% experiments.
\balapinc has two tunable parameters - 1) a threshold 
that indicates the \balapinc score above which all examples are
labeled as positive, 2) the maximum number of features to
consider for each word. We use the tuning set to tune the two
parameters as well as the various hyper-parameters associated with
the models.
%\yv{Should we mention hyperparameter values in the appendix? -- Maybe. How big will the hyperparam table be? I feel it is too much detail and we can just provide the grid over which we search. -- yeah worth skipping in this version, maybe we provide grid in the next version and put exact values on github or something}
\ignore{Preliminary experiments indicated that \balapinc yielded
better results than \slqs. Hence, we use \balapinc as our primary 
entailment score for all experiments, and compare the behavior of the two scores
in the analysis (\S\ref{sec:qual})}

\begin{table*}[htbp]
\begin{threeparttable}
  \begin{minipage}{\textwidth}
    \footnotesize
    \centering
    \begin{tabular}{lL{6.5cm}ccc}
      \toprule
      Language & Parallel Data & \#sent. & Monolingual Data & \#sent.\\
      \midrule
      English & -- & -- & Wackypedia~\cite{baroni2009wacky} & 43M\\
      \midrule
      \multirow{2}{*}{Arabic} & ISI~\cite{arabicparallel} & \multirow{2}{*}{1.1M} & \multirow{2}{*}{Arabic Gigaword 3.0~\cite{arabic-giga2007}} & \multirow{2}{*}{17M}\\
               & NewsCommentary, Wikipedia~\cite{tiedemann2012parallel} &  &  & \\
      \midrule
      Chinese & FBIS (LDC2003E14) & 9.5M & Chinese Gigaword 5.0~\cite{chinese-giga2011} & 58M\\ \midrule
      \multirow{2}{*}{French} & Europarl~\cite{koehn2005europarl} & \multirow{2}{*}{2.7M} & \multirow{2}{*}{Wikipedia$^\clubsuit$} & \multirow{2}{*}{20M}\\ 
               & NewsCommentary$^\blacklozenge$, Wikipedia~\cite{tiedemann2012parallel} &  &  & \\
      \midrule
      Russian & Yandex-1M$^\spadesuit$ & 1.6M & Wikipedia$^\clubsuit$ & 22M\\
      \bottomrule
    \end{tabular}

    \begin{tablenotes}\scriptsize
    \item $\blacklozenge$ = \url{www.statmt.org/wmt15/training-parallel-nc-v10.tgz}, $\clubsuit$ = \url{dumps.wikimedia.org/xxwiki/20161201/}
    \item $\spadesuit$ = \url{translate.yandex.ru/corpus}
    \end{tablenotes}
  \end{minipage}
  \end{threeparttable}
  \caption{Training data statistics for different languages. Note that
    while we use parallel corpora for computing translation
    dictionaries, our approach does not require it, and can work with
    any bilingual
    dictionary.}
    \label{tab:datastats}
        \vspace{-0.1in}
\end{table*}

\subsection{Contrastive Approaches}
\label{sec:contr-appr}
We compare our {\sc BiSparse-Dep}  embeddings with the
following approaches:

\ignore{ \footnote{cross-lingual versions of models from
    \cite{vilnis2014word} and ~\cite{trouillon2016complex} performed
    poorly in preliminary experiments.}
}

\paragraph{{\sc Mono-Dep} (Translation baseline)} For word pair ($p_f$, $q_e$) in
test data, we translate $p_f$ to English using the most common translation
in the translation matrix. Hypernymy is then determined using sparse,
dependency based embeddings in English.

\paragraph{{\sc BiSparse-Lex} (Window context)} Predecessor of the {\sc BiSparse-Dep}
model from our previous work \cite{Vyas2016}. This
model induces sparse, cross-lingual embeddings 
using window based context.

\paragraph{{\sc Bivec+} (Window context)} Our extension of the {\sc Bivec}
model of ~\newcite{luong2015}. {\sc Bivec} generates dense,
cross-lingual embeddings using window based context, by substituting
aligned word pairs within a window in parallel sentences. By default,
{\sc Bivec} only trains using parallel data, and so we initialize
it with monolingually trained window based embeddings to
ensure fair comparison.

% \ignore{
% \xn{not clear what the diff is from your approach} 
%   The dense
%   counterpart of {\sc BiSparse-Dep}. }
\paragraph{{\sc Cl-Dep} (Dependency context)} The model from
~\newcite{vuliceacl2017}, which induces dense, dependency based
cross-lingual embeddings by translating syntactic word-context pairs
using the most common translation, and jointly training a {\tt
  word2vecf}\footnote{\url{bitbucket.org/yoavgo/word2vecf/}} model for
both languages. \newcite{vuliceacl2017} showed improvements for word
similarity and bilingual lexicon induction.  We report the first
results using {\sc Cl-Dep} on this task.

\subsection{Evaluating Robustness of {\sc Bisparse-Dep}}
\label{subsec:setup-lowres}
% Any model that aims to solve a cross-lingual task is likely to run into challenges when exposed to low-resource languages.
We investigate how robust \textsc{BiSparse-Dep}
is when exposed to data scarce settings. Evaluating on a truly low resource language is
complicated by the fact that obtaining an evaluation dataset for such
a language is difficult. Therefore, we simulate such
settings for the languages in our dataset in multiple ways.

\paragraph{No Treebank}
\ignore{As described in \S\ref{sec:depend-embedd-low}, in the absence
of a treebank in a language, a delexicalized parser can be trained
using related languages.}
If a treebank is not available for a language, dependency contexts have to be induced using treebanks from other languages (\S\ref{sec:depend-embedd-low}), which can affect the quality of the dependency-based embeddings.
To simulate this, we train a delexicalized 
parser for the languages in our dataset. We use
treebanks from Slovenian, Ukrainian, Serbian, Polish, Bulgarian,
Slovak and Czech (40k sentences) for training the Russian parser, and treebanks from English, Spanish, German, Portuguese, Swedish and
Italian (66k sentences) for training the French parser. 
UDT does not (yet) have
languages in the same family as Arabic or Chinese, so for the sake of
completeness, we train Arabic and Chinese parsers on delexicalized
treebanks of the language itself. After delexicalized training, the
Labeled Attachment Score (LAS) on the UDT test set dropped by several
points for all languages -- from 76.6\% to 60.0\% for Russian, 83.7\%
to 71.1\% for French, from 76.3\% to 62.4\% for Arabic and from 80.3\%
to 53.3\% for Chinese. The monolingual corpora are then parsed
with these weaker parsers, and co-ocurrences and
dependency contexts are computed as before.

\paragraph{Subsampling Monolingual Data} To simulate low-resource behavior along
another axis, we subsample the monolingual corpora used by {\sc BiSparse-Dep} 
to induce monolingual vectors, $\mathbf{X_{e}},\mathbf{X_{f}}$.
Specifically, we learn $\mathbf{X_{e}}$ and $\mathbf{X_{f}}$ using 
progressively smaller corpora.

\paragraph{Quality of Bilingual Dictionary} We study the impact of
the quality of the bilingual dictionary used to 
create the translation matrix $\mathbf{S}$. This experiment involves
using increasingly smaller parallel corpora to induce the translation dictionary.

% \ignore{
% 	As described in \S~\ref{sec:cross-ling-enta}, our approach first requires
% 	monolingually trained dependency vectors. To train these, we use Arabic Gigaword
% 	v3.0~\cite{arabic-giga2007}, Chinese Gigaword
% 	v5.0~\cite{chinese-giga2011} English Wackypedia~\cite{baroni2009wacky},
% 	and Russian and French Wikipedia
% 	(\url{dumps.wikimedia.org/xxwiki/20161201/}).
	
% 	Our approach also needs a translation matrix for training, which we
% 	obtain from word-alignments derived over parallel corpora.  For En-Fr,
% 	we combined parallel corpora from Europarl~\cite{koehn2005europarl},
% 	News Commentary, and
% 	Wikipedia\footnote{{www.statmt.org/wmt15/training-parallel-nc-v10.tgz}, {sites.google.com/site/iwsltevaluation2015/data-provided}}. For
% 	En-Zh, we use the FBIS parallel corpus (LDC2003E14). For En-Ar, we use
% 	the ISI parallel corpus~\cite{arabicparallel}, combined with corpora
% 	derived from Wikipedia and News
% 	Commentary~\cite{tiedemann2012parallel}. We used the Yandex-1M
% 	(\url{translate.yandex.ru/corpus}) parallel corpus for En-Ru. The
% 	corpus statistics for both monolingual and parallel corpora are shown
% 	in Table~\ref{tab:datastats}.
% }

%%% Local Variables:
%%% mode: latex
%%% TeX-master: "main"
%%% TeX-PDF-mode: t
%%% End: 

\section{Experiments}
\label{sec:exp}
\begin{table*}%[H]
  \footnotesize
  \centering
  \captionsetup[subtable]{position = below}
  \captionsetup[table]{position = below}
  \begin{subtable}{0.5\linewidth}
    \begin{tabular}{l@{ \,}l@{ \,}@{ \,}l@{ \,}@{ \,}l@{ \,}@{ \,}l@{ \,}@{ \,}l@{}}
      \toprule
      % \diagbox[height=0.54cm]{{\small Model}}{{\small En with}} & \multicolumn{1}{l}{Ru} & \multicolumn{1}{l}{Zh}  & \multicolumn{1}{l}{Ar} & \multicolumn{1}{l}{Fr} & Avg.\\ 
      {{Model} {\tiny $\mathbf{\downarrow}$}} {{En With} {\tiny $\mathbf{\rightarrow}$}} & \multicolumn{1}{l}{Ru} & \multicolumn{1}{l}{Zh}  & \multicolumn{1}{l}{Ar} & \multicolumn{1}{l}{Fr} & Avg.\\ 
      \midrule
      \multicolumn{6}{c}{\scriptsize Translation Baseline} \\
      \midrule
      \textsc{Mono-Dep} & 50.1  &  52.3  & 51.8 & 54.5 &  52.2\\
      \midrule
      \multicolumn{6}{c}{\scriptsize Win. Based} \\
      \midrule
      \textsc{BiSparse-Lex} & 56.6 & 53.7 & 50.9 & 52.0 & 53.3\\
      \textsc{Bivec+} & 55.8 & 52.0 & 51.5 & 53.4 & 53.2\\
      \midrule
      \multicolumn{6}{c}{\scriptsize Dep. Based} \\
      \midrule
      \textsc{Cl-Dep} & \bf 60.2 & 54.4 & \bf 56.7* & 53.8 & \bf 56.3 \\ 
      \textsc{BiSparse-Dep} (Full) & 59.0 & 55.9 & 52.6 & 56.6 & 56.0\\
      \textsc{BiSparse-Dep} (Joint) & 53.8 & \bf 57.0* & 52.4 & \bf 59.9* & 55.8\\
      \midrule
      \textsc{BiSparse-Dep} (Unlab) & 55.9 & 51.2 & 53.3 & 55.9 & 54.1\\ 
      \bottomrule
    \end{tabular}
    \caption{Performance on {\sc \hard}.}
    \label{tab:mainexps}
  \end{subtable}%
  \hspace*{0.9em}
  \begin{subtable}{0.5\linewidth}
    \begin{tabular}{l@{ \,}l@{ \,}@{ \,}l@{ \,}@{ \,}l@{ \,}@{ \,}l@{ \,}@{ \,}l@{}}
      \toprule
      % \diagbox[height=0.54cm]{{\small Model}}{{\small En with}} & \multicolumn{1}{l}{Ru} & \multicolumn{1}{l}{Zh}  & \multicolumn{1}{l}{Ar} & \multicolumn{1}{l}{Fr} & Avg.\\ 
      {{Model} {\tiny $\mathbf{\downarrow}$}} {{En With} {\tiny $\mathbf{\rightarrow}$}} & \multicolumn{1}{l}{Ru} & \multicolumn{1}{l}{Zh}  & \multicolumn{1}{l}{Ar} & \multicolumn{1}{l}{Fr} & Avg.\\ 
      \midrule
      \multicolumn{6}{c}{\scriptsize Translation Baseline} \\
      \midrule
      \textsc{Mono-Dep} & 58.7 & 50.0 & 65.1  & 56.9 & 57.7\\
      \midrule
      \multicolumn{6}{c}{\scriptsize Win. Based} \\
      \midrule
      \textsc{BiSparse-Lex} & \textbf{63.8} & 55.8 & 65.8 & 63.2 & 62.2\\
      \textsc{Bivec+}  & 55.9 & 64.9 & 62.2 & 54.1 & 58.3\\
      \midrule
      \multicolumn{6}{c}{\scriptsize Dep. Based} \\
      \midrule
      \textsc{Cl-Dep} & 56.2 & 62.7 & 63.1 & 61.0 & 60.0 \\ 
      \textsc{BiSparse-Dep} (Full)  & 63.6 & \bf 67.3 & \bf 66.8* & \bf 66.7* &	\bf 66.1\\
      \textsc{BiSparse-Dep} (Joint) & 60.6 & 63.6 & 65.9 & 64.9 &	63.8 \\ 
      \midrule
      \textsc{BiSparse-Dep} (Unlab) & 58.6 & 66.7 & 62.4 & 61.5 &	62.4 \\
      % \midrule
      % \textsc{BiSparse-Dep} (Delex)          & 59.4 &		65.7 &	\bf	67.5 &		66.3 &		64.7 \\ 
      \bottomrule
    \end{tabular}
    \caption{Performance on {\sc \easy}.}
    \label{tab:easyexps}
  \end{subtable}
  \caption{Comparing the different approaches from \S\ref{sec:contr-appr} with our {\sc BiSparse-Dep} approach on {\sc \hard} and {\sc \easy} (random baseline= 0.5). {\bf Bold} denotes the best score for each language, and the * on the best score indicates a statistically significant (p $<$ 0.05) improvement over the next best score, using McNemar's test~\cite{mcnemar1947note}. Across both datasets, {\sc BiSparse-Dep} models outperform window based models and the translation baseline on an average.}
  \vspace{-0.1in}
\end{table*}
We aim to answer the following questions -- {\bf (a)} Are dependency based embeddings superior to window based embeddings for identifying cross-lingual hypernymy? (\S\ref{sec:dependency-vs-window}) {\bf (b)} Does directionality in the dependency context help cross-lingual hypernymy identification? (\S\ref{sec:ablat-direct-cont})
{\bf (c)} Are our models robust in data scarce settings (\S\ref{sec:eval-robustn-sc})?
{\bf (d)} Is the answer to {\bf (a)} predicated on the choice of entailment scorer? (\S\ref{sec:choice-enta-score})?

% \begin{figure*}[t]
%   \footnotesize
%   \centering
%   \begin{subfigure}{0.5\textwidth}
%   \end{subfigure}
%   \begin{subfigure}{0.5\textwidth}
%   \end{subfigure}
% \end{figure*}

% \begin{table}[t]
%   \footnotesize
%   \centering
% \end{table}

\subsection{Dependency v/s Window Contexts}
\label{sec:dependency-vs-window}
We compare the performance of models described in
\S\ref{sec:contr-appr} with the \textsc{BiSparse-Dep} ({\sc Full} and
{\sc Joint}) models. We evaluate the models on the two test splits
described in \S\ref{subsec:testsplits} -- {\sc \hard}~ and {\sc
  \easy}.

\paragraph{\hard~ Results}
Table \ref{tab:mainexps} shows the results on {\sc \hard}.  First, the benefit of cross-lingual modeling (as
opposed to translation) is evident in that almost all models (except {\sc Cl-Dep} on French)
outperform the translation baseline. Among dependency based models, {\sc BiSparse-Dep} (\textsc{Full}) and {\sc
  Cl-Dep} consistently outperform both window models, while {\sc BiSparse-Dep} (\textsc{Joint}) outperforms them on all except Russian. \textsc{BiSparse-Dep}
(\textsc{Joint}) is the best model overall for two languages (French and
Chinese), \textsc{Cl-Dep} for one (Arabic), with no statistically significant
differences between {\sc BiSparse-Dep} (\textsc{Joint}) and {\sc Cl-Dep} for Russian.
This confirms that dependency context is more useful than window
context for cross-lingual hypernymy detection.

\paragraph{\easy~ Results}
% \todo{take away should be flipped examples are harder, similar to
%   Vered's finding} 
The trends observed on {\sc \hard~} also hold on {\sc \easy~}
i.e. dependency based models continue to outperform window based models (Table \ref{tab:easyexps}).

Overall, {\sc BiSparse-Dep} (\textsc{Full}) performs best in this setting,
followed closely by {\sc BiSparse-Dep} (\textsc{Joint}). This suggests that the
sibling information encoded in {\sc Joint}  is useful to
distinguish hypernyms from hyponyms ({\sc \hard}~ results), while the dependency labels
encoded in {\sc Full} help to distinguish hypernyms from
co-hyponyms. Also note that all models improve
significantly on the {\sc \easy}~ set, suggesting
that discriminating  hypernyms from
cohyponyms is easier than discriminating them from hyponyms.

% Recall that our
% crowd-sourced hyponyms were generated by annotating flipped hypernym
% pairs, and identifying them requires to detecting asymmetry of the
% hypernymy relation.

While the {\sc BiSparse-Dep} models were generally performing better than window models on both test sets, {\sc Cl-Dep} was not as consistent (e.g., it was worse than the best window model on {\sc \easy}). As shown by \newcite{turney2015experiments}, \balapinc is designed for sparse embeddings and is likely to perform poorly with dense embeddings. This explains the relatively inconsistent performance of {\sc Cl-Dep}.

Besides establishing the challenging nature of our crowd-sourced
set, the experiments on {\sc\easy}~and {\sc\hard}~also demonstrate the ability of the
\textsc{BiSparse-Dep} models to discriminate between different lexical
semantic relations (viz. hypernymy and cohyponymy) in a cross-lingual
setting. We will investigate this ability more carefully in future
work.

\subsection{Ablating Directionality in Context}
\label{sec:ablat-direct-cont}
The context described by the {\sc Full} and {\sc Joint} {\sc BiSparse}
models encodes directional information (\S\ref{sec:depend-based-word-1}) either in the form of label direction ({\sc Full}), or using sibling information ({\sc Joint}).
Does such directionality in the context help to capture the asymmetric
relationship inherent to hypernymy? To answer this, we evaluate a third {\sc
  BiSparse-Dep} model which uses {\sc Unlabeled} dependency contexts. This
is similar to the {\sc Full} context, except we do not concatenate the
label of the relation to the context word (parent or children). For instance, for {\em traveler} in Fig.~\ref{fig:deptree}, contexts will be {\em roamed} and {\em tired}.

Experiments on both {\sc \hard~} and {\sc \easy~}
(bottom row, Tables \ref{tab:mainexps} and \ref{tab:easyexps}) 
highlight that directional information is indeed essential - {\sc Unlabeled}
almost always performs worse than {\sc Full} and {\sc Joint}, and in many cases
worse than even window based models.
% \todo{the next paragraph is not making a strong claim, given last paragraph}
% \ignore{A key point here is that both {\sc Joint} and {\sc Unlabeled} models use an
% unlabeled dependency parse to extract context. Despite this,
% the {\sc Joint} model is consistently better. This
% suggests that if only an unlabeled dependency parse is available in a language, it
% is better to use {\sc Joint} instead of {\sc Unlabeled} context.}

\subsection{Evaluating Robustness of {\sc BiSparse-Dep}}
\label{sec:eval-robustn-sc}
% The experiments in the previous section highlight the robust
% behavior of the {\sc BiSparse} models across languages when 
% exposed to all available information.
% We investigate how the \textsc{BiSparse-Dep} models
% behave when exposed to various data scarce conditions.
% , under three settings:
% {\bf (a)} no dependency treebank in a language 
% {\bf (b)} small monolingual corpus
% {\bf (c)} lower quality bilingual dictionary.

\paragraph{No Treebank}
\begin{table}[t]
  \footnotesize
  \centering
  \begin{tabular}{l@{ \,}l@{ \,}l@{ \,}l@{ \,}l@{ \,}l}
  	\toprule
    % \diagbox[height=0.54cm]{{\small Model}}{{\small En with}} & Ru & Zh & Ar & Fr & Avg.\\
    {{Model} {\tiny $\mathbf{\downarrow}$}} {{En With} {\tiny $\mathbf{\rightarrow}$}} & Ru & Zh & Ar & Fr & Avg.\\
    \midrule
    \multicolumn{6}{c}{\scriptsize \hard} \\
    \midrule
    Best Win. & 56.6 & 53.7 & 51.5 & 53.4 & 53.8 \\
    Delex. & 59.1* & 55.1* & 54.6* & 56.1* & 56.2\\ 
    Best Dep. & 60.2 & 57.0* & 56.7* & 59.9* & 58.5 \\
    \toprule
    \multicolumn{6}{c}{\scriptsize \easy} \\
    \toprule
    Best Win. & 63.8 & 64.9 & 65.8 & 63.2 & 64.4 \\
    Delex. & 59.4 & 65.7* & 67.5* & 66.3* & 64.7 \\
    Best Dep. & 63.6* & 67.3* & 66.8* & 66.7 & 66.1 \\
    \bottomrule
\end{tabular}
\caption{{\bf Robustness in absence of a treebank}: The delexicalized
  model is competitive to the best dependency based and the best
  window based models on both test sets.  For each 
  dataset, * indicates a
  statistically significant (p $<$ 0.05) improvement over the
  next best model in that column, using McNemar's
  test~\cite{mcnemar1947note}.}
\label{tab:delextab}
\end{table}

We run experiments (Table~\ref{tab:delextab}) for all languages with a version of
\textsc{BiSparse-Dep} that use the {\sc Full} context type
for both English and the non-English (target) language, but the target language
contexts are derived from a corpus parsed using a delexicalized
parser (\S\ref{subsec:setup-lowres}).

This model compares favorably on all language pairs against the best
window based and the best dependency based model. In fact, it almost
consistently outperforms the best window based model by several points, and
is only slightly worse than the best dependency-based model.

Further analysis revealed that the good performance of the
delexicalized model is due to the relative robustness of the
delexicalized parser on frequent contexts in the co-occurrence
matrix. Specifically, we found that in French and Russian, the most frequent
contexts were derived from {\tt amod}, {\tt nmod}, {\tt nsubj} and
{\tt dobj} edges.\footnote{Together they make up at least 70\% of the contexts 
%	\yv{For every language? If yes, say so.}
	.} 
For instance, the {\tt nmod} edge appears in 44\%
of Russian contexts and 33\% of the French contexts. The delexicalized
parser predicts both the label and direction of the {\tt nmod} edge
correctly with an F1 of 68.6 for Russian and 69.6 for French. In
contrast, a fully-trained parser achieves a F1 of 76.7 for Russian and
76.8 for French for the same edge.

% \subsection{Size of Bilingual Dictionary}

\paragraph{Small Monolingual Corpus}
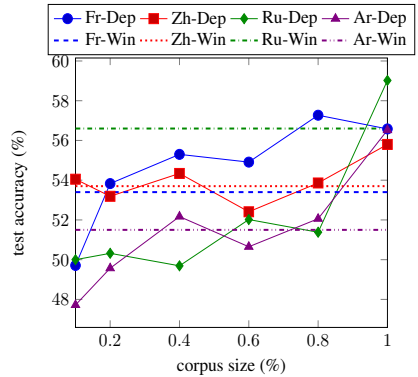
\begin{figure}[t]
  \centering
  \resizebox{5.5cm}{5cm}{
    \begin{tikzpicture} % plotting with data in columns
      \begin{axis}[
        xmin=0.1,
        xmax=1.0,
        xlabel=corpus size (\%),
        ylabel=test accuracy (\%),
        % extra y ticks={56.6,53.4,51.5,53.7},
        % extra y tick style={grid=major,
        %   yticklabels={Ru-Win,Fr-Win,Ar-Win,Zh-Win},
        %   yticklabel style={yshift=1ex, anchor=west, font=\footnotesize,rotate=-30}},
        legend style={
          at={(0.5,1.2)},       % 0.5 times xmin,xmax and -0.18 times ymin,ymax, think a$
          anchor=north,
          legend columns=4       % no columns, only rows
      },
    ]
    \addplot table [x=frac, y=fr, col sep=comma] {size_tuning.txt};
    \addplot table [x=frac, y=zh, col sep=comma] {size_tuning.txt};
    % increasing 54 to increase green-ness
    \addplot[mark=diamond*,color=green!54!black] table [x=frac, y=ru, col sep=comma] {size_tuning.txt};
    \addplot[mark=triangle*,color=violet] table [x=frac, y=ar, col sep=comma] {size_tuning.txt};
    % \addplot[mark=*] coordinates {(0,1)};
    % \node[label={0:{Ru-Win}}] at (axis cs: 0,56.6) {};
    \addplot[mark=None,dashed,very thick,color=blue] table [x=frac, y=frwin, col sep=comma] {size_tuning.txt};
    \addplot[mark=None,dotted,very thick,color=red] table [x=frac, y=zhwin, col sep=comma] {size_tuning.txt};
    \addplot[mark=None,dashdotted,very thick,color=green!54!black] table [x=frac, y=ruwin, col sep=comma] {size_tuning.txt};
    \addplot[mark=None,dashdotdotted,very thick,color=violet] table [x=frac, y=arwin, col sep=comma] {size_tuning.txt};
    \legend{Fr-Dep,Zh-Dep,Ru-Dep,Ar-Dep,Fr-Win,Zh-Win,Ru-Win,Ar-Win}
      \end{axis}
\end{tikzpicture}
  }
  \caption{{\bf Robustness to Small Corpus} For most languages, {\sc BiSparse-Dep}
    outperforms the corresponding best window based model for each
    language on {\sc \hard}, with about 40\% of the monolingual corpora.}
  \label{fig:wtune}
\end{figure}

%%% Local Variables:
%%% mode: latex
%%% TeX-master: "main"
%%% TeX-PDF-mode: t
%%% End: 

In Figure
\ref{fig:wtune}, we use increasingly smaller monolingual corpora (10\%, 20\%,
40\%, 60\% and 80\%) sampled at random to
induce the monolingual vectors for \textsc{BiSparse-Dep} ({\sc Full}) model. Trends (Figure \ref{fig:wtune}) indicate that
\textsc{BisSparse-Dep} models that use only 40\% of the original data
remain competitive with the \textsc{BisSparse-Lex} model that has
access to the full data. Robust performance with smaller
monolingual corpora is helpful since large-enough monolingual corpora
are not always easily available.

\paragraph{Quality of Bilingual Dictionary}
\begin{figure}[t]
  \centering
  \resizebox{5.5cm}{5cm}{
    \begin{tikzpicture} % plotting with data in columns
      \begin{axis}[
        xmin=0.1,
        xmax=1.0,
        xlabel=frac. of parallel corpus,
        ylabel=test accuracy (\%),
        % extra y ticks={56.6,53.4,51.5},
        % extra y tick style={grid=major,
        %   yticklabels={Ru-Win,Fr-Win,Ar-Win},
        %   yticklabel style={yshift=1ex, anchor=west, font=\footnotesize}},
        legend style={
          at={(0.5,1.2)},       % 0.5 times xmin,xmax and -0.18 times ymin,ymax, think a$
          anchor=north,
          legend columns=4       % no columns, only rows
        },
        ]
        \addplot table [x=frac, y=fr, col sep=comma] {tune.csv};
        \addplot table [x=frac, y=zh, col sep=comma] {tune.csv};
        \addplot[mark=diamond*,color=green!54!black] table [x=frac, y=ru, col sep=comma] {tune.csv};
        \addplot[mark=triangle*,color=violet] table [x=frac, y=ar, col sep=comma] {tune.csv};
        \addplot[mark=None,dashed,very thick,color=blue] table [x=frac, y=frwin, col sep=comma] {size_tuning.txt};
        \addplot[mark=None,dotted,very thick,color=red] table [x=frac, y=zhwin, col sep=comma] {size_tuning.txt};
        \addplot[mark=None,dashdotted,very thick,color=green!54!black] table [x=frac, y=ruwin, col sep=comma] {size_tuning.txt};
        \addplot[mark=None,dashdotdotted,very thick,color=violet] table [x=frac, y=arwin, col sep=comma] {size_tuning.txt};
%         \legend{Ar-Dep,Fr-Dep,Ru-Dep,Zh-Dep,Ar-Win,Fr-Win,Ru-Win,Zh-Win}
        \legend{Fr-Dep,Zh-Dep,Ru-Dep,Ar-Dep,Fr-Win,Zh-Win,Ru-Win,Ar-Win}
      \end{axis}
    \end{tikzpicture}
  }
  \caption{{\bf Robustness to Noisy Dictionary} For most languages, {\sc BiSparse-Dep}
    outperforms the corresponding best window based model on {\sc \hard}, with increasingly lower quality dictionaries.}
  \label{fig:wtune}
\end{figure}
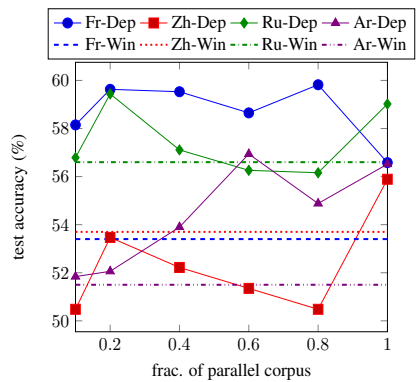

%%% Local Variables:
%%% mode: latex
%%% TeX-master: "main"
%%% TeX-PDF-mode: t
%%% End: 

% It is prudent to analyze how big a bilingual dictionary is needed as
% the score matrix $\mathbf{S}$ by our approach to get good results.
Bilingual dictionaries derived from smaller amounts of parallel data are likely to be of lower quality than those derived from larger corpora. Hence, to analyze the impact of dictionary quality on \textsc{BiSparse-Dep} (\textsc{Full}), 
we use increasingly smaller 
parallel corpora to induce bilingual dictionaries used as the
score matrix $\mathbf{S}$ (\S\ref{subsec:bisparse}).  We use the top 10\%, 20\%, 40\%, 
60\%  and 80\% sentences from the parallel corpora. The trends in Figure \ref{fig:wtune}
show that even with a lower quality dictionary,
\textsc{BiSparse-Dep} performs better than \textsc{BiSparse-Lex}.
% This reinforces our model's ability to perform well in low resource
% settings along another dimension.

\subsection{Choice of Entailment Scorer}
\label{sec:choice-enta-score}
We change the entailment scorer from \balapinc to
\slqs\cite{santus2014chasing} and redo experiments from
\S\ref{sec:dependency-vs-window} to see if the
conclusions drawn depend on the choice of the entailment
scorer.
\slqs is based on the distributional
informativeness hypothesis, which states that hypernyms are less
``informative'' than hyponyms, because they occur in more general
contexts. The informativeness $E_u$ of a word $u$ is defined to be the
median entropy of its top $N$ dimensions,
$E_u = median_{k=1}^N H(c_k)$, where $H(c_i)$ denotes the entropy of
dimension $c_i$. The \slqs score for a pair $(u,v)$ is the relative
difference in entropies,
\begin{align*}
SLQS(u \rightarrow v) = 1 - \frac{E_u}{E_v}
\end{align*}
%\xn{just realized this is the first equation in the paper!}
% Details on computing \slqs~ scores can be found in Appendix (\S\ref{sec:computing-slqs-score}).
Recent
work~\cite{shwartz2017hypernymy} has found \slqs to be more successful
than other metrics in monolingual hypernymy detection.

The trends observed in these experiments are consistent 
with those in \S\ref{sec:dependency-vs-window} -- both {\sc BiSparse-Dep} models still outperform window-based
models. Also, the delexicalized version of {\sc BiSparse-Dep} 
outperforms the window-based models, showing that the robust behavior
demonstrated in \S\ref{sec:eval-robustn-sc}
is also invariant across metrics. 

We also found that using \balapinc led to better results than \slqs.
% Using the {\sc BiSparse-Dep} (Full) model with \balapinc performed
% better than the {\sc BiSparse-Dep} (Full) model with \slqs in three
% out of the four languages (all but Ar).
For both {\sc BiSparse-Dep} models, \balapinc
wins across the board for two languages (Russian and Chinese), and wins half the
time for the other two languages compared to \slqs. We leave detailed comparison of 
these and other scores to future work.
% This validates our decision to use \balapinc as the primary
% metric, but at the same time, leaves the door open for further investigations
% into the use of \slqs for cross-lingual hypernymy.

%%% Local Variables:
%%% mode: latex
%%% TeX-master: "main"
%%% TeX-PDF-mode: t
%%% End: 

% \section{Analysis}
% \label{sec:qual}
% \input{qual}

\section{Conclusion}
We introduced \textsc{BiSparse-Dep}, a new distributional approach for
identifying cross-lingual hypernymy, based on cross-lingual embeddings
derived from dependency contexts. We showed that using {\sc
  BiSparse-Dep} is superior for the cross-lingual hypernymy detection task,
when compared to standard window based models and a translation
baseline. Further analysis also showed that {\sc BiSparse-Dep} is
robust to various low-resource settings.  In principle, {\sc
  BiSparse-Dep} can be used for any language that has a bilingual
dictionary with English and a ``related'' language with a treebank. We
also introduced crowd-sourced cross-lingual hypernymy datasets for
four languages for future evaluations.
 
Our approach has the potential to complement existing work on creating
cross-lingual ontologies such as BabelNet and the Open Multilingual Wordnet, which are noisy because they are compiled
semi-automatically, and have limited language coverage. In general, distributional
approaches can help refine ontology construction for any language
where sufficient resources are available.

\ignore{\xn{Not clear how this paragraph logically connects, and how the conclusion flows.
You could say "we introduced a new distributional approach for identifying cross-lingual hypernymy."
The summarize the results (2nd paraphgraph)
Then pop back up one level and discuss bigger picture implications:
"Our approach has the potential to complement existing work on creating cross-lingual ontologies, which are noisy because they are compiled semi-automatically, and have limited language coverage. Distributional approaches can help refine ontology construction for any language where sufficient resources are available."
And then finally turn to future work.
"It remains to be seen how our approach performs for other language pairs beyond simluated low-resource settings. We anticipate that replacing our delexicalized parser with more sophisticated transfer strategies (cite) might be beneficial in such settings." }}

% Distributional approaches for identifying cross-lingual hypernymy are
% complementary to work on creating cross-lingual ontologies which are
% compiled semi-automatically, and have limited language coverage. In
% contrast, distributional approaches can, in principle, be trained for
% any language.

% Furthermore, such approaches are easy to use and provide a graded
% score, instead of absolute links found by graph traversal.

% ALREADY SAID ABOVE and showed our approach consistently outperforms
% other baselines that use translation and window-based embeddings.

It remains to be seen how our approach performs for other language pairs beyond simluated low-resource settings. We anticipate that replacing our delexicalized parser with more sophisticated transfer strategies~\cite{rasooliTACL,aufrant-wisniewski-yvon2016COLING} might be beneficial in such settings.While our delexicalized parsing based approach exhibits robustness, it can benefit from more
sophisticated approaches for transfer parsing~\cite{rasooliTACL,aufrant-wisniewski-yvon2016COLING} to improve
parser performance.  \ignore{For instance, the lexical feature gap in
delexicalized parser can be bridged by using cross-lingual embeddings
tailored for dependency parsing~\cite{guo2016distributed} or using
multilingual clusters~\cite{tackstrom2012cross}. The robust nature of
our approach also encourages applications to truly low-resource
settings such as taxonomy induction or extension of existing
taxonomies.} We aim to explore these and other directions in the
future.

% A natural future direction is to transfer supervised hypernymy
% discovery~\cite{Roller2014,espinosa2016supervised} approaches
% cross-lingually for this task.  The ability to perform well in low
% resource settings also suggests we can induce taxonomies for new
% languages, or even extend existing taxonomies.  Our approach for low
% resource languages can also benefit from more sophisticated
% approaches for transfer parsing, like selecting related languages in
% a more principled
% way~\cite{rasooliTACL,aufrant-wisniewski-yvon2016COLING} to improve
% parser performance.  For instance, the lexical feature gap in
% delexicalized parser can be bridged by using cross-lingual
% embeddings tailored for dependency parsing~\cite{guo2016distributed}
% or using multilingual clusters~\cite{tackstrom2012cross}.

\ignore{
\todo{Future direction -- contextual lexical entailment like Vered, cross-lingual taxonomy contruction using this method, DONE supervised x-ling entailment by faeture selection}
}
%%% Local Variables:
%%% mode: latex
%%% TeX-master: "main"
%%% TeX-PDF-mode: t
%%% End: 

\section*{Acknowledgments}
The authors would like to thank the members of the CLIP lab at the
University of Maryland, members of the Cognitive Computation Group at
the University of Pennsylvania, and the anonymous reviewers from
EMNLP/CoNLL 2017 and NAACL 2018 for their constructive feedback. YV
and MC were funded in part by research awards from Amazon, Google, and
the Clare Boothe Luce Foundation. SU and DR were supported by
Contract HR0011-15-2-0025 with the US Defense Advanced Research
Projects Agency (DARPA).

\bibliography{references,cited,ref_yogi}
\bibliographystyle{acl_natbib}
\appendix

\end{document}